\documentclass{article}

\usepackage[nonatbib,final]{neurips_2023}

\usepackage[utf8]{inputenc} 
\usepackage[T1]{fontenc}    
\usepackage{hyperref}       
\usepackage{url}            
\usepackage{booktabs}       
\usepackage{amsfonts}       
\usepackage{nicefrac}       
\usepackage{microtype}      
\usepackage{xcolor}         
\usepackage{multirow}
\usepackage{graphicx}
\usepackage{authblk}
\usepackage{makecell}
\usepackage{wrapfig}

\title{Planting a SEED of Vision in Large Language Model}

\begin{document}

\author{
\textbf{Yuying Ge$^{1\star}$ \qquad Yixiao Ge$^{1,2\star \dagger}$ \qquad Ziyun Zeng$^{2}$ \qquad Xintao Wang$^{1,2}$ \qquad Ying Shan$^{1,2}$} 

$^{1}$Tencent AI Lab \qquad $^{2}$ARC Lab, Tencent PCG

\url{https://github.com/AILab-CVC/SEED}
}
 \renewcommand{\thefootnote}{\fnsymbol{footnote}}
 		\footnotetext[1]{Equal Contribution.} 
   \footnotetext[2]{Correspondence to \texttt{yixiaoge@tencent.com}.}

\maketitle

\begin{abstract}

We present \textbf{SEED}, an elaborate image tokenizer that empowers Large Language Models (LLMs) with the emergent ability to \textbf{SEE} and \textbf{D}raw at the same time.
Research on image tokenizers has previously reached an impasse, as frameworks employing quantized visual tokens have lost prominence due to subpar performance and convergence in multimodal comprehension (compared to BLIP-2, etc.) or generation (compared to Stable Diffusion, etc.).
Despite the limitations, we remain confident in its natural capacity to unify visual and textual representations, facilitating scalable multimodal training with LLM's original recipe.
In this study, we identify two crucial principles for the architecture and training of SEED that effectively ease subsequent alignment with LLMs.
(1) Image tokens should be independent of 2D physical patch positions and instead be produced with a \textit{1D causal dependency}, exhibiting intrinsic interdependence that aligns with the left-to-right autoregressive prediction mechanism in LLMs.
(2) Image tokens should capture \textit{high-level semantics} consistent with the degree of semantic abstraction in words, and be optimized for both discriminativeness and reconstruction during the tokenizer training phase.
As a result, the off-the-shelf LLM is able to perform both image-to-text and text-to-image generation by incorporating our SEED through efficient LoRA tuning.
Comprehensive multimodal pretraining and instruction tuning, which may yield improved results, are reserved for future investigation.
This version of SEED was trained in 5.7 days using only 64 V100 GPUs and 5M publicly available image-text pairs.
Our preliminary study emphasizes the great potential of discrete visual tokens in versatile multimodal LLMs and the importance of proper image tokenizers in broader research.
\end{abstract}

\section{Introduction}
In recent years, Large Language Models~\cite{touvron2023llama, brown2020language, chowdhery2022palm} (LLMs) pre-trained on massive text corpus with straightforward training objectives such as next-word prediction have exhibited remarkable abilities to understand, reason, and generate texts across a variety of open-ended tasks. %
Recent studies further exploit the strong generality of LLMs to improve visual understanding or generation tasks, collectively referred to as Multimodal LLM (MLLM).
For example, previous work~\cite{ye2023mplug, li2023blip, zhu2023minigpt, gao2023llama, liu2023visual} perform open-ended visual QAs through aligning visual features of a pre-trained image encoder (e.g., CLIP-ViT) with the input embedding space of LLMs.
GILL~\cite{koh2023gill} empowers LLM with the image generation ability by aligning its output embedding space with the pre-trained Stable Diffusion (SD) model~\cite{rombach2022high}.

While these studies have contributed to technological advancements, MLLMs have yet to achieve the remarkable success of LLMs in terms of emergent capabilities.
We have made a bold assumption that the premise for the emergence of multimodal capabilities is that text and images can be represented and processed \textbf{interchangeably} in a unified autoregressive Transformer.
Fortunately, we have just found consensus in concurrent works \cite{sun2023generative,yu2023scaling}, all employing image-to-text and text-to-image generation tasks to demonstrate the emergent ability of unifying visual comprehension and generation in one framework.
Regardless of discrete or continuous visual tokens, the training paradigm can be summarised into three stages: \textbf{visual tokenizer training, multimodal pretraining, and multimodal instruction tuning}.
While concurrent studies primarily emphasize multimodal training (the latter two stages), this work focuses more on the visual tokenizer (the first stage).

We posit that a proper visual tokenizer can facilitate the follow-up multimodal training by (i) easing the semantic alignment between visual and word tokens, and (ii) enabling LLM's original training recipe (i.e., next-word prediction) for multimodal data without specific adaptation for visual tokens.
Representing images as a sequence of discrete IDs is naturally compatible with the autoregressive training objective of LLMs.
But unfortunately, works \cite{ramesh2021zero,ding2021cogview} that utilize discretized visual tokens for multimodal tasks have receded from prominence, as such models generally rely on super-scale training to converge, leading to substantial training costs.
Moreover, we empirically found that the dominant tokenizer VQ-VAE~\cite{van2017neural} in existing works captures too low-level information for LLMs to effectively perform multimodal comprehension tasks.
Existing image tokenizers fail to meet the requirements of unifying visual understanding/generation tasks and facilitating multimodal training.

\begin{figure}
	\centering
	\includegraphics[width=1.0\linewidth]{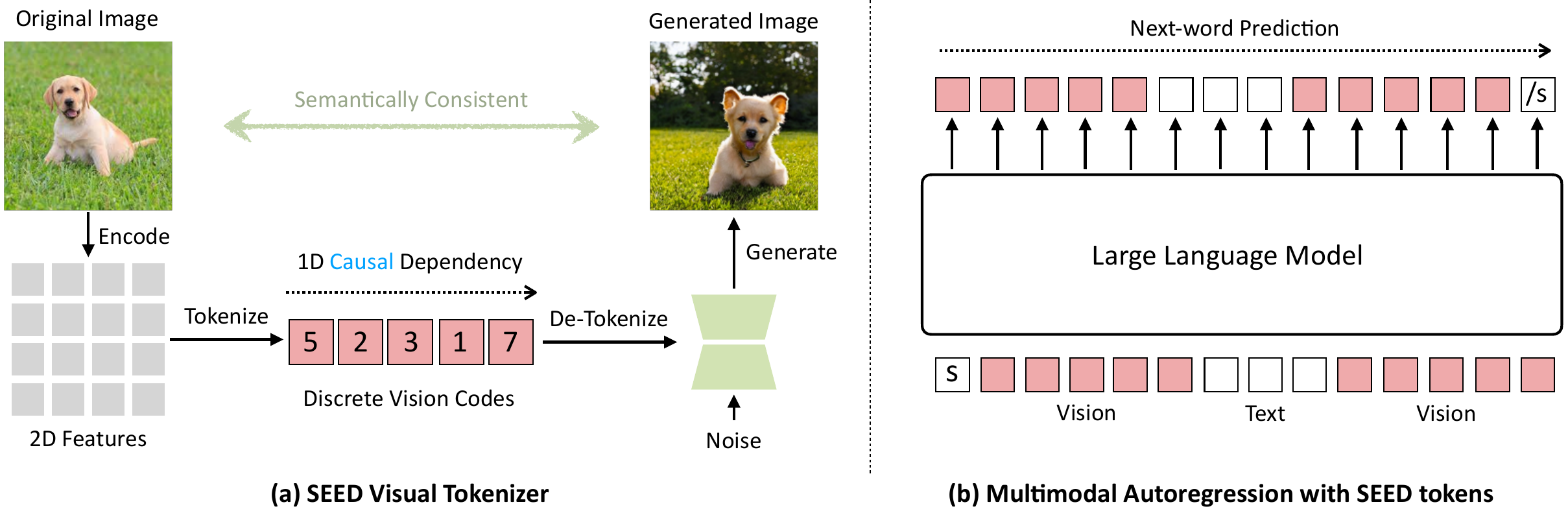}
	\caption{(a) The proposed SEED is a discrete image tokenizer, producing quantized visual codes with 1D causal dependency and high-level semantics. (b) SEED visual tokens enable LLMs to perform both visual comprehension and generation through multimodal autoregression with interleaved image-text data.}
	\label{fig:intro}
\end{figure}

To this end, we introduce \textbf{SEED}, a VQ-based image tokenizer that produces discrete visual codes with 1D causal dependency and necessary high-level semantics for both visual comprehension and generation tasks, as shown in Fig. \ref{fig:intro}. 
The off-the-shelf LLMs can be readily equipped with SEED by treating discrete visual tokens as new words and updating the vocabulary with mapped visual codes.
In the paper, we present an MLLM by tuning the pre-trained LLM with low-rank adaptation (LoRA) to efficiently align with the SEED tokenizer.

We would like to emphasize the design principles of SEED.
(1) \textit{Why causal-dependent tokens?}
Existing visual tokens (e.g., from VQ-VAE or CLIP-ViT) are generated using 2D context, which is incompatible with the unidirectional attention in dominant LLMs and counterintuitive for text-to-image tasks requiring raster order prediction. Thus, we convert 2D raster-ordered embeddings into a sequence of semantic codes with 1D causal dependency.
(2) \textit{Why high-level semantics?}
Since visual and textual tokens in LLMs are expected to be interoperable—sharing weights and training objectives—they should encompass the same degree of semantics to prevent misalignment, i.e., the high-level semantics inherently present in words.\footnote{
While focusing on high-level semantics during tokenization, it is still possible to achieve accurate spatial structural control, such as layout and mask conditions, in image generation tasks. These spatial structural prompts can be tokenized similarly, as demonstrated by the success of SD \cite{rombach2022high,li2023gligen}.}

Specifically, the SEED tokenizer is composed of a ViT encoder, Causal Q-Former, VQ Codebook, Reverse Q-Former, and a UNet decoder.
The ViT encoder and UNet decoder are directly derived from the pre-trained BLIP-2 and SD models, respectively.
(1) \textit{Tokenize:}
Causal Q-Former converts 2D raster-ordered features produced by the ViT encoder into a sequence of causal semantic embeddings, which are further discretized by the VQ Codebook.
(2) \textit{De-Tokenize:}
The discrete visual codes are decoded into generation embeddings via Reverse Q-Former. The generation embeddings are aligned with the latent space of SD so that realistic images with consistent semantics can be generated using the off-the-shelf SD-UNet.

During SEED training, only Causal Q-Former, VQ Codebook, and Reverse Q-Former are tunable.
Causal Q-Former is optimized by image-text contrastive loss.
VQ Codebook and Reverse Q-Former are trained toward the objectives of dual reconstruction, i.e., the reconstruction between continuous causal embeddings and discrete causal codes, the reconstruction between generation embeddings and the paired textual features.
The training objectives ensure that SEED encapsulates the essential semantics for both visual comprehension and generation.
Quantitative results indicate that discrete SEED tokens exhibit competitive performance in text-image retrieval compared to BLIP-2, and in image generation compared to Stable Diffusion.
With further multimodal autoregressive training, SEED-OPT$_{\rm{2.7B}}$ (efficiently tuned via LoRA using 5M image-text pairs) effectively performs image-to-text and text-to-image tasks, yielding promising results in zero-shot image captioning and visual QA, as well as generating high-quality images.

This effort aims to integrate multimodal comprehension and generation tasks within an LLM using discrete visual tokens. Our initial exploration of proper tokenizer designs strives to promote the development of emergent multimodal capabilities.
Future work can further scale up training for a better tokenizer and leverage stronger LLMs (e.g., LLaMA~\cite{touvron2023llama}) for comprehensive multimodal pretraining and instruction tuning.

\section{SEED Visual Tokenizer}

\subsection{Pilot Experiments of Baseline Tokenizers} 

Visual tokenizer aims to represent the image as a sequence of discrete tokens. Previous work~\cite{van2017neural,ramesh2021zero,esser2021taming} trains a Vector Quantized Variational AutoEncoders (VQ-VAE) by reconstructing image pixels, while Beit v2~\cite{peng2022beit} propose vector-quantized knowledge distillation (VQ-KD) to train a visual tokenizer by reconstructing high-level features from the teacher model. We conduct two experiments to respectively align discrete representations of VQ-VAE and Beit v2 with OPT$_{\rm{2.7B}}$~\cite{zhang2022opt} model on CC3M~\cite{sharma2018conceptual} dataset. We evaluate the performance with zero-shot image captioning on COCO~\cite{lin2014microsoft}. VQ-VAE achieves CIDEr 34.0 while Beit v2 achieves 42.0. The experiment results demonstrate that a high-level visual tokenizer, which captures semantic representations of images instead of low-level image details is more effective for multimodal comprehension.

\subsection{Architecture}

\begin{figure}
	\centering
	\includegraphics[width=1.0\linewidth]{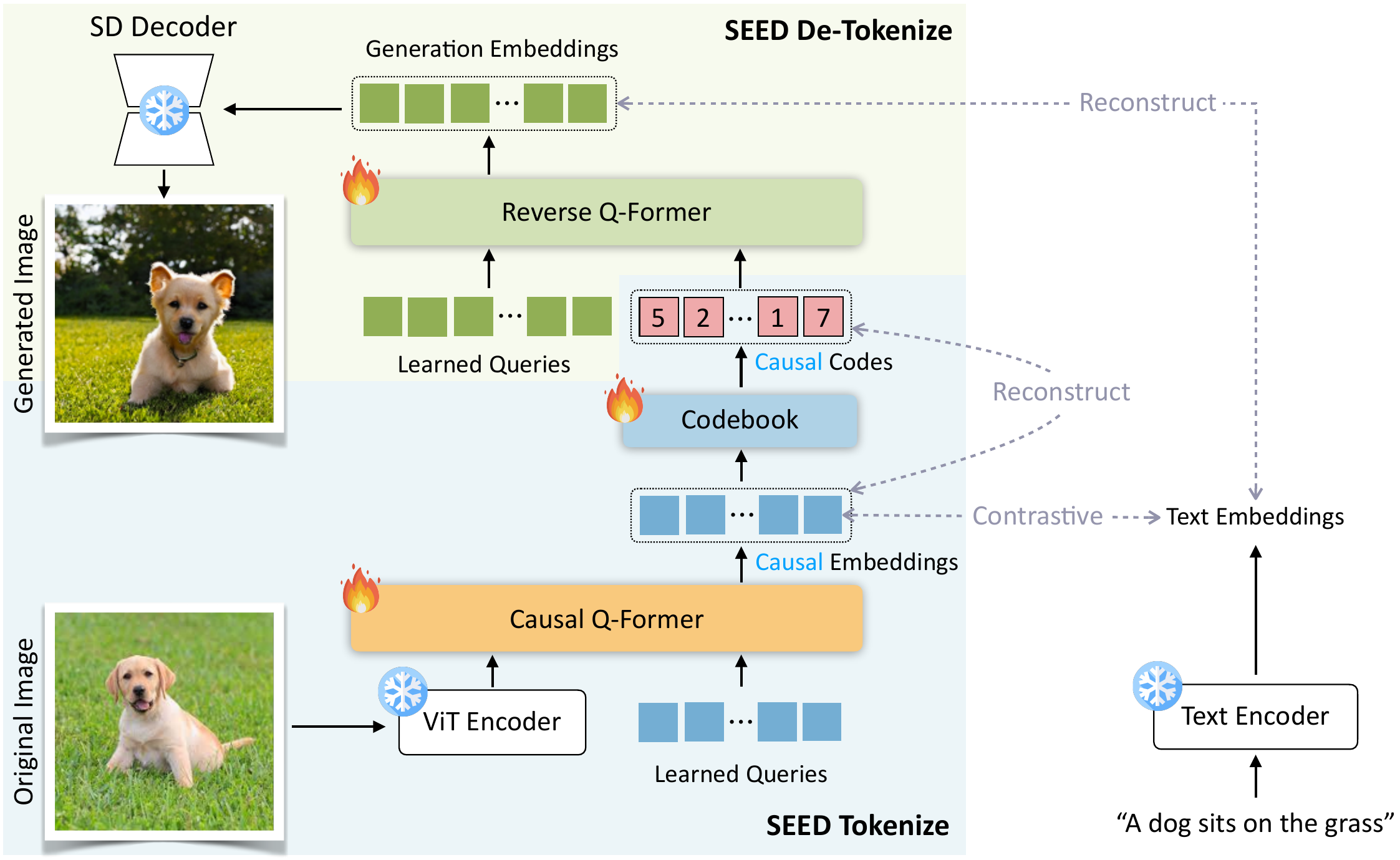}
 
\caption{Overview of our \textbf{SEED} tokenizer, which produces discrete visual codes with causal dependency and high-level semantics.}
	\label{fig:tokenizer}
\end{figure}

In this work, we introduce a VQ-based image tokenizer \textbf{SEED} to produce discrete visual codes with 1D causal dependency and high-level semantics. Specifically, as shown in Fig.~\ref{fig:tokenizer}, the SEED tokenizer is composed of a ViT image encoder~\cite{dosovitskiy2020image}, Causal Q-Former, VQ Codebook, Reverse Q-Former, and a UNet decoder~\cite{rombach2022high}. The ViT encoder and UNet decoder are directly derived
from the pre-trained BLIP-2 and SD models, respectively. We first train a Causal Q-Former to convert 2D raster-ordered features (16$\times$16 tokens) produced by the ViT encoder into a sequence of causal semantic embeddings (32 tokens). We then train a visual codebook to discretize the causal embeddings to quantized visual codes (32 tokens) with causal dependency. We employ a Reverse Q-Former to decode the visual codes into generation embeddings (77 tokens), which are aligned with the latent space of the pre-trained Stable Diffusion (SD) model.

\subsubsection{Training Stage I: Causal Q-Former}

As shown in Fig.~\ref{fig:tokenizer}, a set number of learnable query embeddings (32 tokens) and features of a pre-trained ViT image encoder are fed into the Causal Q-former to encode a fixed number of causal embeddings (32 tokens) of the input image. Specifically, the query embeddings can interact with only previous queries through self-attention layers with causal mask, and interact with frozen image features through cross-attention layers. We adopt contrastive learning to optimize Causal Q-former fine-tuned from pre-trained BLIP-2 Q-Former on 5M image-text pairs including CC3M~\cite{sharma2018conceptual}, Unsplash~\cite{Unsplash}, and COCO dataset~\cite{lin2014microsoft}. We use contrastive loss to maximize the similarity between the \textbf{final} causal embedding and text features of the corresponding caption, while minimizing the similarity between the \textbf{final} causal embedding and text features of other captions in a batch. 

{\flushleft \bf Evaluation of Causal Embeddings.} We evaluate the performance of Causal Q-Former on the zero-shot image-text retrieval task using \textbf{COCO}~\cite{lin2014microsoft} and  \textbf{Flickr30K}~\cite{young2014image} dataset following BLIP-2. The performance is measured by \emph{Recall@K} (R@K) for both image-to-text retrieval and text-to-image retrieval. Note that we adopt the dual-stream paradigm for inference and remove the image-txt-matching (ITM) rerank module in BLIP-2 for a fair comparison. As shown in Tab.~\ref{tab:retrieval}, our Causal Q-former achieves better results than BLIP-2 in terms of an aggregated metric \emph{Recall@mean}. It demonstrates that the output query embeddings with causal dependency do not drop performance than the output embeddings with bi-directional attention in BLIP-2. 

\begin{table}[t]
\centering
\caption{Evaluation of zero-shot Image-Text Retrieval. Causal codes are quantized causal embeddings.}
\resizebox{1.\columnwidth}{!}{
\begin{tabular}{lccccccc|ccccccc}
\toprule
\multirow{3}{*}{Model} & \multicolumn{7}{c|}{Flickr30K (1K test set)}                 & \multicolumn{7}{c}{COCO (5K test set)}                      \\
                                                      & \multicolumn{3}{c}{Image $\rightarrow$ Text} & \multicolumn{3}{c}{Text $\rightarrow$ Image} & &\multicolumn{3}{c}{Image $\rightarrow$ Text} & \multicolumn{3}{c}{Text $\rightarrow$ Image}& \\
\cmidrule(l){2-15}
                                                         & R@1        & R@5      & R@10      & R@1       & R@5       & R@10   &R@mean   & R@1       & R@5       & R@10      & R@1       & R@5       & R@10 &R@mean     \\
\midrule
BLIP-2 ~\cite{li2023blip}                                     & 81.9       & 98.4     & 99.7      & 82.4      & 96.5      & 98.4     &92.9 & 65.3      & 89.9      & 95.3      & 59.1      & 82.7      & 89.4  &80.3   \\
SEED (causal emb)    &90.0	&99.6	&99.9	&80.0	&95.3	&97.6	&93.7	&71.9	&91.1	&95.9	&56.7	&80.7	&87.7 &80.7\\
SEED (causal code) &86.3	&98.6	&99.5	&75.9	&93.2	&96.7	&91.7	&65.7	&88.1	&93.8	&52.5	&78.0	&86.0	&77.4\\
\bottomrule
\label{tab:retrieval}
\vspace{-25pt}
\end{tabular}}
\end{table}

\subsubsection{Training Stage II: Visual Quantization and De-tokenization}

As shown in Fig.~\ref{fig:tokenizer}, we train a VQ codebook to discretize the causal embeddings (32 tokens) into quantized visual codes (32 tokens) on 5M image-text pairs including CC3M, Unsplash, and COCO dataset. Specifically, a quantizer looks up the nearest neighbor in the codebook for each causal embedding and obtains the corresponding code. We employ a decoder, which is a multi-layer Transformer~\cite{dosovitskiy2020image}, to reconstruct the continuous causal embeddings from discrete codes. During training, we maximize the cosine similarity between the output of the decoder and the causal embeddings. We further employ a Reverse Q-Former to reconstruct the textual features of a frozen stable diffusion model from discrete codes. A set number of learnable query embeddings (77 tokens) are fed into the Reverse Q-Former. The query embeddings interact with each other through self-attention layers, and interact with causal codes (32 tokens) through cross-attention layers for the output generation embeddings (77 tokens). During training, we minimize the MSE loss between generation embeddings and text features of SD. During inference, the generation embeddings can be fed into the SD-UNet to decode realistic images.

{\flushleft \bf Evaluation of Causal Codes.} We evaluate the performance of SEED tokenizer on zero-shot image-text retrieval, where the reconstructed causal embeddings from causal codes are used for retrieval. 
As shown in Tab.~\ref{tab:retrieval}, discrete SEED tokens exhibit competitive performance compared to BLIP-2. 

\begin{wraptable}{r}{5.5cm}
\centering
\vspace{-10pt}
\caption{Evaluation of Image Generation with CLIP similarity as the metric. }
\resizebox{.3\columnwidth}{!}{
\begin{tabular}{lcc}
\toprule
Model & COCO  & Flickr30K \\
\midrule
GILL~\cite{koh2023gill}  & 67.45 & 65.16     \\
SD~\cite{rombach2022high}    & 68.43 & 65.40      \\
SEED    & 68.23 & 65.22      \\
\bottomrule
\label{tab:clip_score}
\vspace{-10pt}
\end{tabular}}
\end{wraptable}
We further evaluate image generation on \textbf{COCO} and \textbf{Flickr30K} dataset. SEED first discretizes input images into causal codes (32 tokens) and obtain generation embeddings (77 tokens) from Reverse Q-Former, which are fed into the SD-UNet for the reconstructed images. For the baseline model GILL~\cite{koh2023generating} and SD~\cite{rombach2022high}, images are generated from corresponding captions of the input images. We follow GILL~\cite{koh2023generating} to compute the CLIP similarity as the evaluation metric for benchmarking the semantic consistency. As shown in Tab.~\ref{tab:clip_score}, compared with the upper bound SD, our SEED only slightly drops performance, and outperforms GILL in image generation.

\begin{figure}
	\centering
	\includegraphics[width=0.86\linewidth]{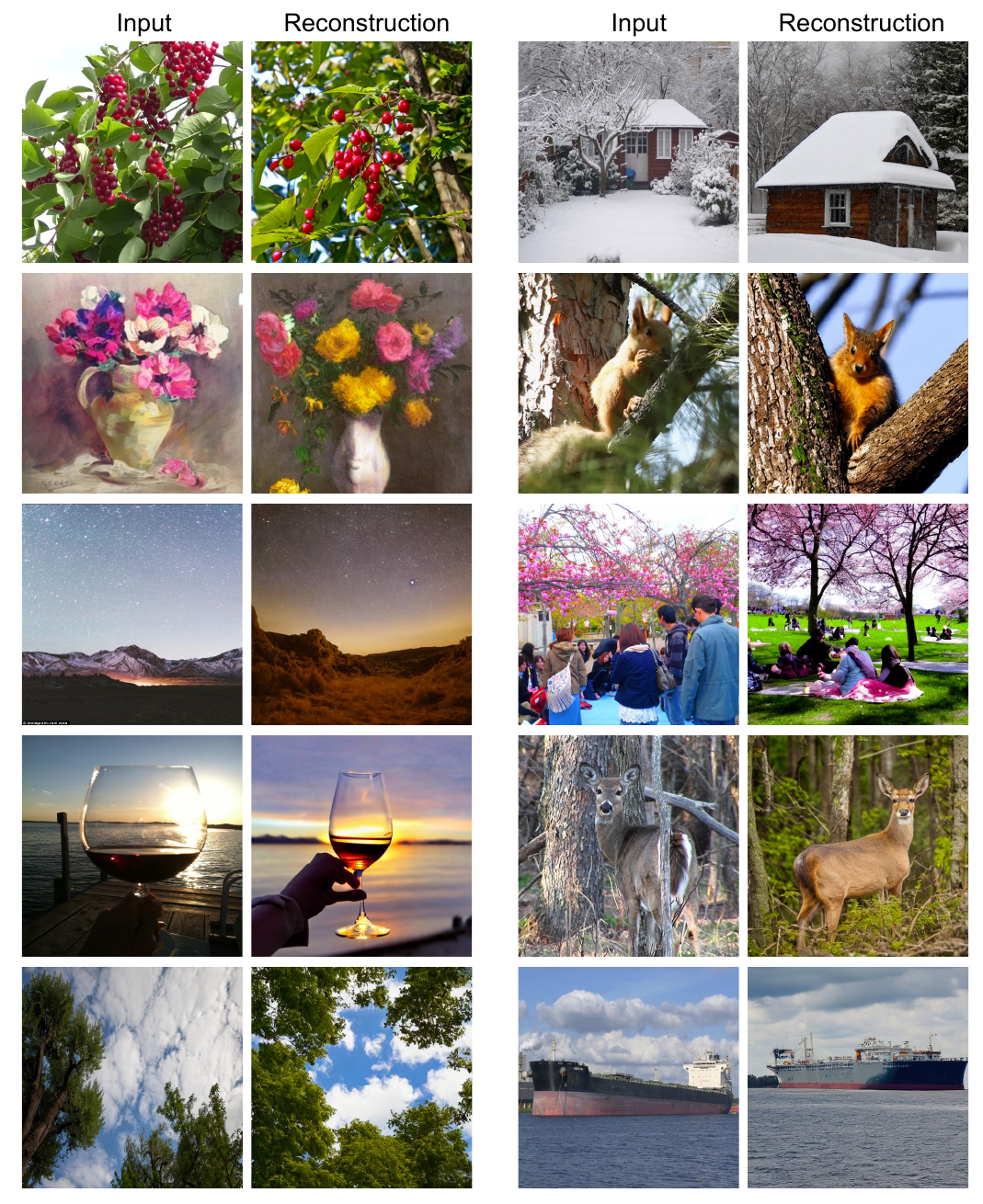}
  	\vspace{-10pt}	
	\caption{Reconstruction images of SEED tokenizer (i.e., original image $\rightarrow$ SEED tokenize $\rightarrow$ causal visual codes $\rightarrow$ SEED de-tokenize $\rightarrow$ reconstructed image), which are semantically consistent with the original input images.}
	\vspace{-15pt}	
	\label{fig:reconstruction}
\end{figure}

{\flushleft \bf Visualization of Reconstructed Images.} We visualize the reconstructed images of SEED in Fig.~\ref{fig:reconstruction}. Through utilizing the Reverse Q-Former to obtain the generation embeddings from the causal visual codes of the input image, realistic images can be generated using the off-the-shelf SD-UNet, which maintain consistent semantics with input images.

\textit{The above evaluation and visualization demonstrate the versatility of SEED visual tokens for both comprehension and generation tasks.}

\section{Multimodal Autoregression with SEED Visual Tokens}
\begin{figure}
	\centering
	\includegraphics[width=1.0\linewidth]{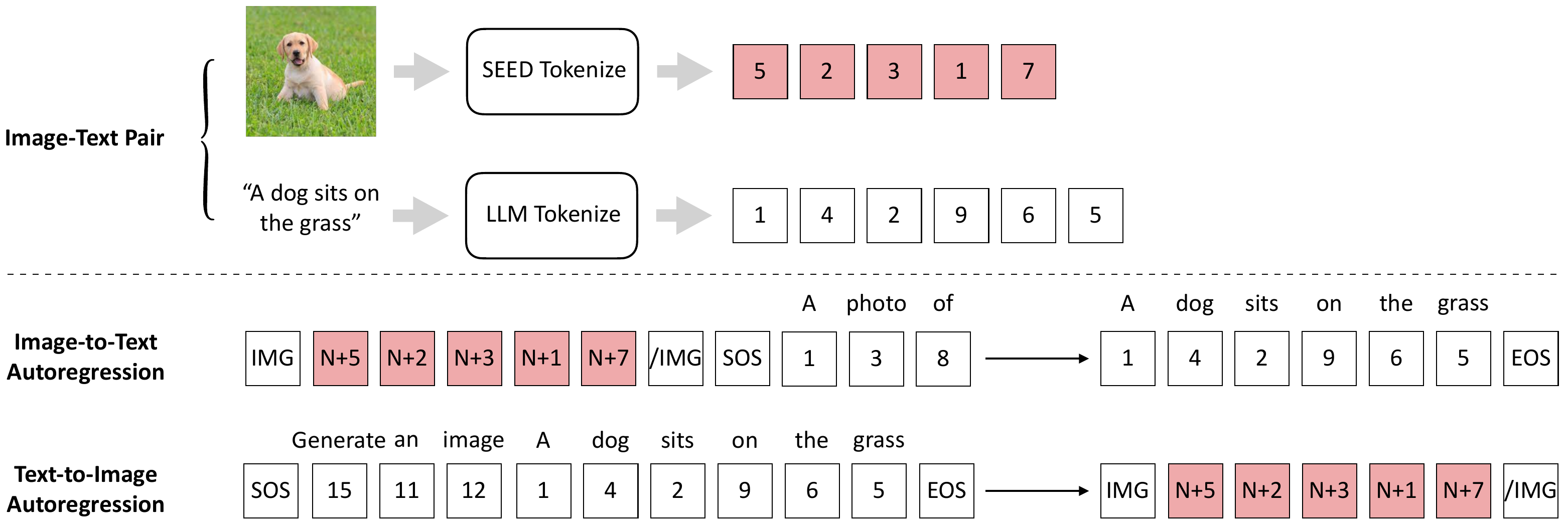}
	\caption{Overview of the multimodal autoregressive training for SEED-OPT$_{\rm{2.7B}}$ using efficient LoRA tuning. It was trained in 44 hours using only 64 V100 GPUs and 5M image-caption pairs.}
	\label{fig:autoregression}
\end{figure}

Based on the pre-trained SEED tokenizer, we present SEED-OPT$_{\rm{2.7B}}$ through fine-tuning a low-rank adaption (LoRA)~\ module on a OPT$_{\rm{2.7B}}$~\cite{zhang2022opt} model with 5M image-text pairs including CC3M, Unsplash and COCO dataset. As shown in Fig.~\ref{fig:autoregression}, we perform image-to-text and text-to-image autoregressive pre-training for unified multimodal comprehension and generation. 

{\flushleft \bf Image-to-Text Autoregression.} We first perform image-to-text autoregression to align the vocabulary of the pre-trained VQ codebook with OPT$_{\rm{2.7B}}$. Specifically, we use a fully-connected (FC) layer to linearly project the causal codes from the visual tokenizer into the same dimension as the word embeddings of OPT$_{\rm{2.7B}}$. The projected causal codes and the word embeddings of the prefix ``A photo of'' are concatenated as the input of the OPT$_{\rm{2.7B}}$. The text tokens of the corresponding caption is used as the generation target. We freeze OPT$_{\rm{2.7B}}$ and fine-tune LoRA with the training objective of predicting the next text token.

{\flushleft \bf Text-to-Image Autoregression.} We then jointly perform image-to-text and text-to-image autoregression to empower the LLM with the ability to generate vision tokens in addition to text tokens. For text-to-image autoregressive pre-training, the word embeddings of the prefix ``Generate an image'' and a caption are fed into OPT$_{\rm{2.7B}}$. The visual codes of the corresponding image from our pre-trained tokenizer are used as the generation target. We freeze OPT$_{\rm{2.7B}}$ and fine-tune LoRA with the training objective of predicting the next vision token.

During inference, given the prompt ``Generate an image'' and a text description, SEED-OPT$_{\rm{2.7B}}$ predicts the visual tokens autoregressively. The output visual tokens are fed into the Reverse Q-Former for generation embeddings, which can be decoded to generate a realistic image via SD-UNet.

\begin{table}[t]
\centering
\caption{Comparison between BLIP-2 (pre-trained with 129M image-text pairs) and SEED-OPT$_{\rm{2.7B}}$ (5M pairs) on zero-shot Image Captioning and Visual Question Answering. S: SPICE, M: METEOR, R: ROUGE\textsubscript{L}, B: BLEU, C: CIDEr.}
\resizebox{0.9\columnwidth}{!}{
\begin{tabular}{lcccc|ccccc|cc}
\toprule
\multirow{3}{*}{Models}               & \multicolumn{4}{c|}{NoCaps}                                                                          & \multicolumn{5}{c|}{COCO}      &VQAv2 &GQA \\
\cmidrule(l){2-12}
                                        &                                  \multicolumn{1}{c}{in} & \multicolumn{1}{c}{near} & \multicolumn{1}{c}{out} & \multicolumn{1}{c|}{overall} & \multicolumn{5}{c|}{Karpathy test}
                                        &&\\
& S & S & S  & S   &B@4 &M &R   & C     & S    &Top-1 &Top-1 \\
\midrule
BLIP-2 OPT\textsubscript{2.7B}~\cite{li2023blip}            &14.4	&13.8	&13.4	&13.8	&39.7	&28.9	&59.3	&131.0	&22.9 &51.9 &32.6  \\
SEED-OPT$_{\rm{2.7B}}$ &12.5	&12.3	&12.2	&12.3	&34.6	&28.4	&56.4	&119.0	&22.0 &42.8&28.8\\
\bottomrule
\label{tab:vqa}
	\vspace{-5pt}	
\end{tabular}}
\end{table}

{\flushleft \bf Evaluation of Multimodal Understanding.} We evaluate the performance of SEED-OPT$_{\rm{2.7B}}$ with zero-shot image captioning and visual question answering (vqa). For image captioning, we evaluate on both \textbf{COCO}~\cite{lin2014microsoft} test set and \textbf{NoCaps}~\cite{agrawal2019nocaps} validation set and report BLEU@K (B@K), METEOR (M), ROUGE\textsubscript{L} (R), CIDEr (C), and SPICE (S) with the prompt ``a photo of''. For visual question answering,  we evaluate on \textbf{VQAv2}~\cite{goyal2017making} validation set and \textbf{GQA}~\cite{gqa} test set and report Top-1 accuracy with the prompt ``Question: \{\} Short answer.'' As shown in Tab.~\ref{tab:vqa}, compared with BLIP-2, which are trained on \textbf{129M} image-text pairs, our SEED-OPT$_{\rm{2.7B}}$ trained on \textbf{5M} pairs achieves promising results on zero-shot image captioning and visual question answering with SEED discrete visual tokens. Note that different from concurrent work CM3Leon~\cite{yu2023scaling} that uses image captioning and vqa datasets for \textbf{supervised fine-tuning}, our SEED-OPT$_{\rm{2.7B}}$ pre-trained with image-to-text autoregression using the prefix ``A photo of'' can perform zero-shot visual question answering by understanding free-form questions and predicting open-form answers. 

\begin{figure}
	\centering
	\includegraphics[width=1.0\linewidth]{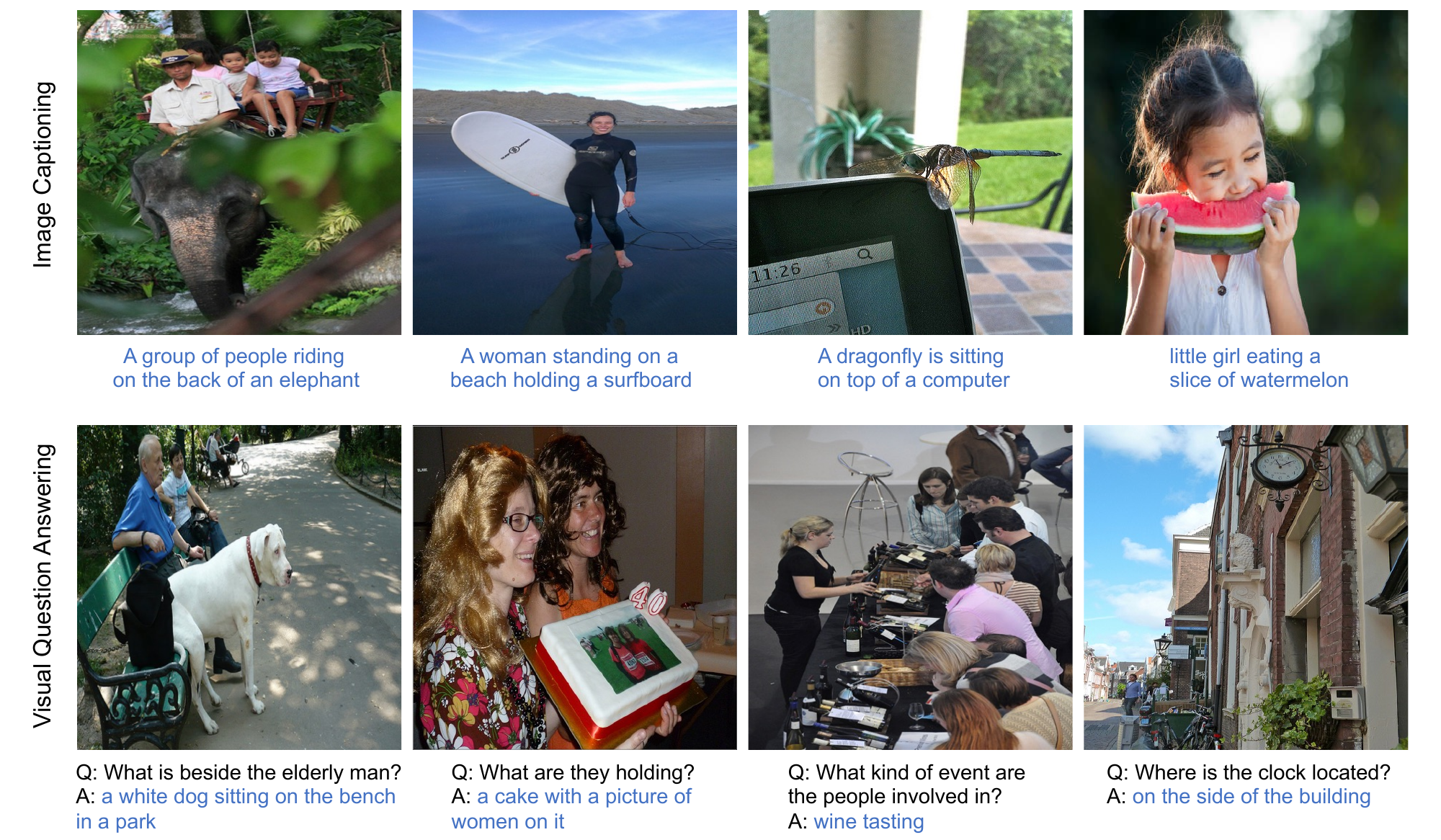}
  	\vspace{-15pt}	
	\caption{Qualitative examples of SEED-OPT$_{\rm{2.7B}}$ on image captioning (with a prompt ``a photo of'') and open-ended visual question answering. Our model has not been trained on any VQA dataset.}
	\vspace{-10pt}	
	\label{fig:vqa}
\end{figure}

We also show qualitative examples of SEED-OPT$_{\rm{2.7B}}$ on image captioning (with a prompt ``a photo of'') and vqa. As shown in Fig.~\ref{fig:vqa}, our model can generate captions than describe the visual content, and answer a variety of questions.

\begin{figure}
	\centering
	\includegraphics[width=1.0\linewidth]{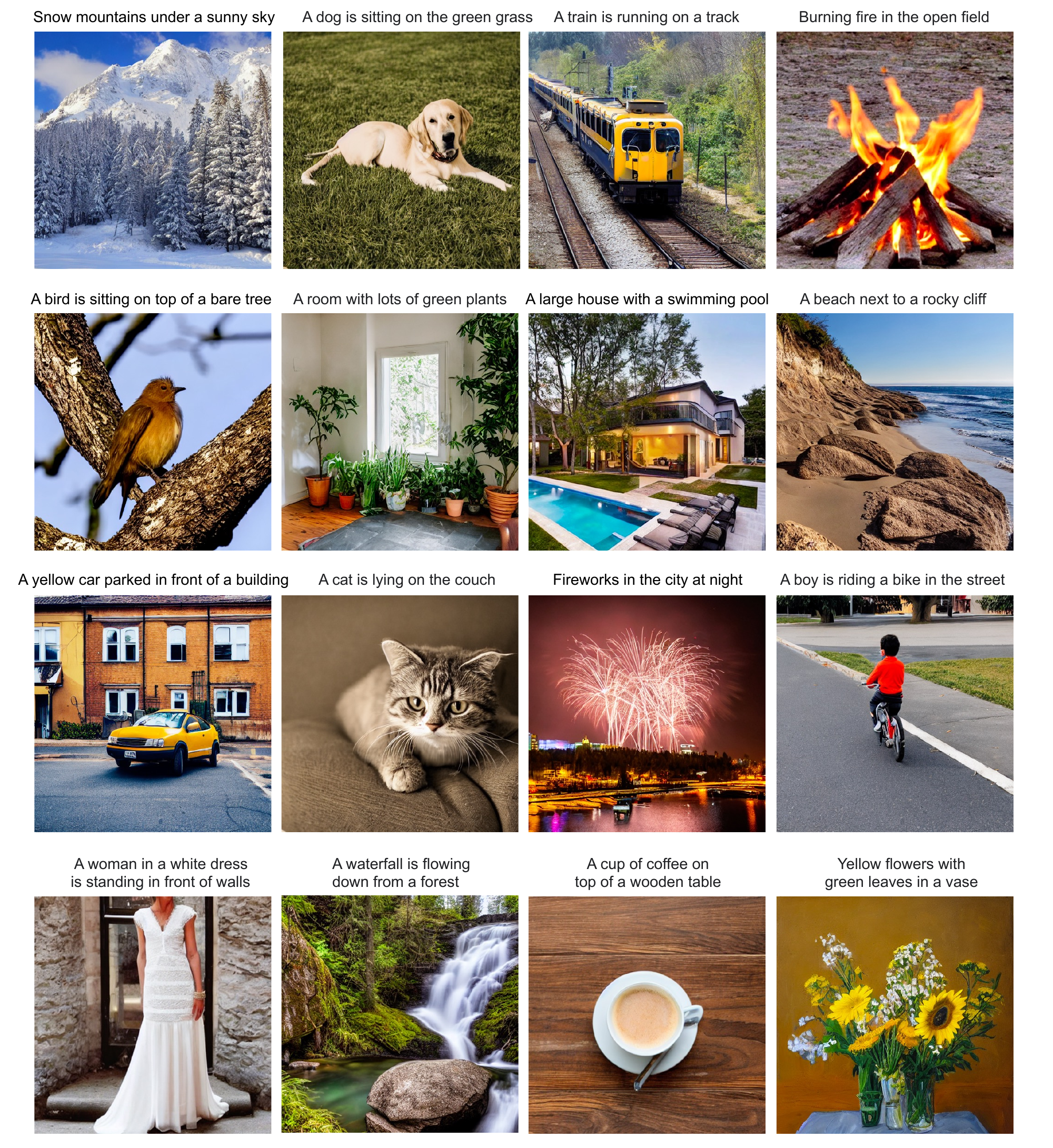}
  	\vspace{-15pt}	
	\caption{Text-to-image generation results when inferring with SEED-OPT$_{\rm{2.7B}}$.}
	\vspace{-10pt}	
	\label{fig:text-to-image}
\end{figure}

{\flushleft \bf Evaluation of Multimodal Generation.} We showcase qualitative examples of text-to-image generation results with our SEED-OPT$_{\rm{2.7B}}$ in Fig.~\ref{fig:text-to-image}. Given the textual description, SEED-OPT$_{\rm{2.7B}}$ can generate realistic images that are semantically relevant to the description.

\textit{SEED can facilitate alignment between visual tokens and LLMs, as evidenced by SEED-OPT$_{\rm{2.7B}}$, already capable of performing text-to-image and image-to-text generation tasks after LoRA tuning.}

\section{Related Work}\label{sec:related_work}

{\flushleft \bf Multimodal Large Language Models for Comprehension.} 
With the impressive success of Large language models~\cite{touvron2023llama, brown2020language, chowdhery2022palm} (LLMs), recent studies work on Multimodal LLM (MLLM) to improve visual comprehension through utilizing the strong generality of LLMs. Previous work~\cite{ye2023mplug,li2023blip,zhu2023minigpt,zhang2023llama,gao2023llama,liu2023visual,alayrac2022flamingo,driess2023palm} align visual features of pre-trained image encoder with LLMs on image-text datasets, and empower LLMs with the ability to interpret visual information with textual descriptions. However, these work commonly use the prediction of the next text token as the training objective and exert no supervision for vision data, thus can only output texts given multimodal vision and language inputs. 

{\flushleft \bf Multimodal Large Language Models for Generation.} 
To empower LLMs with the image generation ability, CogView~\cite{ding2021cogview} pre-trains a visual tokenizer by reconstructing image pixels, and fine-tunes GPT models~\cite{brown2020language, radford2019language} with the objective of next token prediction, where both image and text tokens are equally treated. GILL~\cite{koh2023generating} learns a mapping between the embeddings of a LLM and a frozen pretrained image generation model. Both work aim to generate images with LLMs, without being explicitly designed for multimodal comprehension.

{\flushleft \bf Visual Tokenizer.} Visual tokenizer aims to represent the image as a sequence of discrete tokens similar to natural language. Previous work~\cite{van2017neural,ramesh2021zero,esser2021taming} trains a Vector Quantized Variational AutoEncoders (VQ-VAE) as a visual tokenizer by reconstructing the pixels of the input images, which captures only low-level details of images such as color, texture and edge. Beit v2~\cite{peng2022beit} trains a semantic-rich visual tokenizer through reconstructing high-level features from the teacher model, but its visual codes from 2D features of a vision transformer~\cite{dosovitskiy2020image} are incompatible with the unidirectional attention in dominant LLMs for multimodal generation.

\section{Conclusion}
We present SEED, a discrete image tokenizer, designed based on the premise that visual tokens compatible with LLMs should capture high-level semantics while being generated with a 1D causal dependency.
SEED enables LLMs to be trained with multimodal data following the original recipe of text (i.e., next-word prediction), which is mature and scalable. The trained multimodal LLM is capable of both image-to-text and text-to-image generation tasks, taking one more step toward emergent multimodal capabilities.
We hope that our SEED would draw increased attention to visual tokenizers. A more rational visual tokenizer could substantially reduce the cost and complexity of multimodal LLM training, promoting lower-carbon, large-scale model training. Moreover, we eagerly anticipate the ``germination'' of vision (imagination) seeds within LLMs.
The project is still in progress. Stay tuned for more updates!

\subsection*{Acknowledgements}
We sincerely acknowledge Sijie Zhao (Tencent AI Lab) and Chen Li (ARC Lab, Tencent PCG) for their engaging discussions.

{\small
\bibliographystyle{unsrt}
\bibliography{SEED}
}

\end{document}